\documentclass[journal]{IEEEtai}
\usepackage[colorlinks,urlcolor=blue,linkcolor=blue,citecolor=blue]{hyperref}
\usepackage{color,array}
\usepackage{graphicx}
\usepackage{booktabs}
\usepackage{amsmath}
\usepackage{hyperref}
\usepackage[ruled,vlined]{algorithm2e}
\usepackage[backend=bibtex]{biblatex}
\bibliography{A_decision-making_framework_for_recommended_maintenance_of_road_segment}

\ifCLASSOPTIONcompsoc \usepackage[caption=false,font=normalsize,label on
t=sf,textfont=sf]{subfig}
\else
\usepackage[caption=false,font=footnotesize]{subfi g}
\fi

\begin{document}
  
\title{A decision-making framework for recommended maintenance of road segments}

\author{\IEEEauthorblockN{Haoyu~Sun, Yan Yan}~\\
\IEEEauthorblockA{Collelge~of~Computing~Artificial~Intelligence\\
Illinois~Institute~of~Technology~Chicago,~Illinois~60616\\
Email:~hsun29@hawk.iit.edu, yyan34@.iit.edu}}

\maketitle

\begin{abstract}
Due to limited budgets allocated for road maintenance projects in various countries, road management departments face difficulties in making scientific maintenance decisions. This paper aims to provide road management departments with more scientific decision tools and evidence. The framework proposed in this paper mainly has the following four innovative points: 1) Predicting pavement performance deterioration levels of road sections as decision basis rather than accurately predicting specific indicator values; 2) Determining maintenance route priorities based on multiple factors; 3) Making maintenance plan decisions by establishing deep reinforcement learning models to formulate predictive strategies based on past maintenance performance evaluations, while considering both technical and management indicators; 4) Determining repair section priorities according to actual and suggested repair effects. By resolving these four issues, the framework can make intelligent decisions regarding optimal maintenance plans and sections, taking into account limited funds and historical maintenance management experiences.

\end{abstract}

\begin{IEEEImpStatement}
Deciding which road segments are prior ones is very important in road maintenance because of a limited budget, based on literature research, the most widely applied method is to predict the performance index of pavement. However, recent customer feedback shows that this kind of forecasting method does not meet management needs. With a significant increase in user satisfaction after adopting our decision-making framework, the technology is ready to support users in a wide variety of applications including road management departments and construction companies, consulting companies, etc.
\end{IEEEImpStatement}


\section{Introduction}
\IEEEPARstart{R}{oad} maintenance is critical to transportation infrastructure management. With the increasing volume of traffic and the deterioration of road conditions, road maintenance has become a challenging task for road management departments worldwide. Limited budgets for road maintenance make it difficult for road management departments to keep up with maintenance needs\cite{MLPPPs}. Furthermore, the effectiveness of maintenance techniques and materials can vary significantly between different construction companies. Thus, selecting the right maintenance process, materials, and construction company is critical for obtaining the highest maintenance effect within budget constraints\cite{MLPPPs, priority}.

In recent years, advancements in infrastructure construction capabilities and road maintenance techniques have led to the need for an effective decision-making framework for road maintenance. However, most road management departments still lack a reliable and efficient method for making informed decisions about road maintenance. The reasons are as follows:
\begin{enumerate}
\item[1)]A large amount of historical maintenance data has not been applied and analyzed. Most of these data are kept as work records; no data mining and analysis work has been done. 

\item[2)]As a traditional industry, there is little mastery and application of big data analysis technology or artificial intelligence technology, however, traditional analysis tools and methods\cite{Fuzzyreg4pvmt} cannot analyze and process the sudden increase in data in recent years.

\item[3)]As we all know, the decline in pavement performance is the result of the combined effects of various complex factors\cite{LTPP}. At the same time, because the detection of pavement performance has not achieved a long-term and continuous monitoring mechanism, it is difficult to achieve accurate pavement performance prediction in the absence of data from other factors. Therefore, it is impossible to identify the road segments that need to be prioritized for maintenance. 

\end{enumerate}

This paper proposes a decision-making framework for recommended maintenance of road segments to address this issue. The framework utilizes reinforcement learning, a form of artificial intelligence, to analyze historical maintenance data and evaluate different materials and maintenance processes. The framework also incorporates road surface disease prediction to predict future maintenance needs and optimize maintenance planning, the framework we proposed contains several following parts.

\begin{enumerate}
\item[1)]Determine the road surface performance level through the distribution of regional road network data and construct a prediction model to predict the performance level of the pavement in the next year. 

\item[2)]Based on the historical data of the number of pavement diseases and structure diseases in the evaluation unit (10m) construct a structure diseases prediction model.

\item[3)]We propose a reinforcement learning (RL)\cite{RLintran} algorithm based on the evaluation of road maintenance historical records, combining the output results of 1) and step 2) build data structure adapted to the RL algorithm and output the final recommend maintenance of road segments.

\end{enumerate}

This paper presents the development and implementation of the proposed framework and evaluates its performance through experiments and engineering evaluations. The results demonstrate that the framework provides a reliable and efficient method for decision-making in road maintenance, enabling road management departments to make informed decisions that maximize the effectiveness of maintenance within the available budget, at the same time, this framework also can link the project-level and network-level decisions\cite{MLPPPs}.

\section{Related work}
The related work on the performance of pavement prediction is in section \ref{PPP}, and the reinforcement learning (RL) will be reviewed in \ref{LR}.

\subsection{Performance of Pavement Prediction}\label{PPP}
Predicting the performance of pavement is an older topic, this research has been studied in depth by researchers a long time ago, and the most widely used methods so far are nothing more than three types: traditional regression model\cite{Fuzzyreg4pvmt}, fuzzy regression\cite{Fuzzyreg4pvmt}, and machine learning\cite{pfmnspvmntpre, MLPPPs,  MLprepfmspvmt}.
In general, the traditional regression model is easy to understand, but the traditional linear regression model only considers the time series but ignores the other factor which also influences the performance of the pavement. 

Since the simple linear regression model has large errors in engineering applications, scholar Nang-Fei Pan et al proposed a method of fuzzy regression\cite{Fuzzyreg4pvmt, fuzzyLRanal}. In Pan's research, the authors add more influence factors to the proposed model. Machine learning models are the most popular in road maintenance in recent years, including random forest\cite{MLprepfmspvmt}, Markov chain\cite{Markovppp}, and ANN\cite{ATTOHOKINE1999291} et al. Various studies have shown that in pavement performance prediction under the action of multiple factors, several methods of machine learning can show excellent performance. 

However, regardless of the prediction method, scholars pursue accurate prediction indicators in order to determine the timing of maintenance based on the predicted indicators. But in this paper, we creatively propose fuzzy prediction of deterioration levels as one metric for the maintenance plan recommendation model.

\subsection{Reinforcement Learning}\label{LR}
Reinforcement learning (RL) is extremely critical in uncertain decision-making tasks. The application field not only exists in data mining\cite{RLinDataMing}  and robotics\cite{RLinRobotic} but has also played an extremely critical role in computer vision tasks\cite{RLinCV} in recent years. Similarly, in the field of highway maintenance, a small number of scholars have begun to use reinforcement learning for scientific decision-making. For instance, Yao and Zhen have published research papers on the utilization of RL in highway maintenance in 2022\cite{RLinHighway}. In addition, Fan and Zhang published research to use the RL model for resilient road network recovery under earthquake or flooding hazards\cite{RLinrdrc}. They have achieved more effective results through experiments conducted on selected road sections. Its powerful decision-making ability makes reinforcement learning a core of the framework we proposed. Based on reinforcement learning theory, given a data set including the possible states and the probability of output of each state $[p_1, S_1,p_2, S_2,...,p_n, S_n]$, $\pi(S)$ is the state recommended by the policy. The expected utility of the set is:
\begin{equation}
EU([p_1, S_1,p_2, S_2,...,p_n, S_n])=\Sigma_{i=1}^n P_iU(S_i)
\end{equation}
The artificial decision-making agent will choose the maximum expected utility value as the next action $a$.
In the real world, the agent not only chooses the action but also needs to find the best strategy, thus the Bellman Equation has proposed to solve this problem that can be described:

\begin{equation}
U^\pi (S) = \Sigma _{S'}P(S'|S,\pi(S))[R(S,\pi(S),S')+\gamma U^\pi (S')]
\end{equation}

where $R(S,\pi(S), S')$ denotes the reward the agent receives when it transitions from state $S$ to state $S'$ via policy $\pi(S)$. 
The best strategy which will be selected by the agent can be defined as:

\begin{equation}
U(S) =\max_{a\in A(S)} \Sigma _{S'}P(S'|S,a)[R(S,a,S')+\gamma U(S')]
\end{equation}

Bellman Optimality Equation is similar to Bellman Equation, $A(s)$ denotes a possible action set.
Constructing a simulator requires a very deep understanding of the scene. Therefore, constructing the simulator itself may be a more difficult problem than making decisions and making predictions\cite{PPP}. This is a limitation of reinforcement learning but it is also a problem that needs to be addressed in this study.

\section{Proposed framework}

\begin{figure*}[!t]
    \centering
    \includegraphics[width=4.5in]{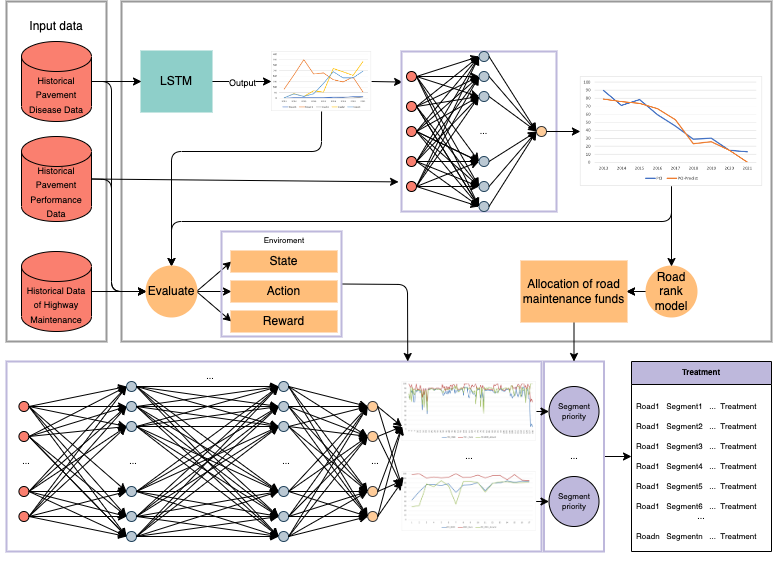}
    \caption{Flowchart of our proposal, composed of three modules for the pavement performance prediction model, the priority ranking model, and the maintenance road segment recommendation model.}
    \label{fig: flowchart}
\end{figure*}

The proposed decision framework consists of three modules: pavement performance prediction model, maintenance section recommendation, and maintenance priority ranking model. Figure \ref{fig: flowchart} illustrates the flowchart of our proposed approach. The first step involves utilizing two models to perform prediction tasks. The prediction task involves forecasting the development of various types of pavement distress quantities and future pavement performance using a historical dataset spanning ten years. The prediction results are then combined with maintenance history data and other fundamental road information to construct the environment and reward function required for reinforcement learning. This facilitates the generation of optimal decision recommendations. Finally, the priority ranking model is invoked once again, taking into account the maintenance budget as a constraint, to determine the final set of maintenance sections.

More details about the pavement performance prediction model will introduction in section \ref{pavement performance prediction model}, and maintenance road segment recommendation model in \ref{maintenance road segment recommendation model}, the \ref{RM} part main introduction how is the rank model work.

\subsection{pavement performance prediction model} \label{pavement performance prediction model}

Over the past period up until now, pavement performance prediction has directly determined the annual maintenance plans of highway administration departments. Therefore, it has always been an important topic in the field of highway maintenance, but also an unsolved difficult problem. Many scholars have done a lot of work on pavement performance prediction, but their results are limited to ideal data in the laboratory, because in real engineering applications, due to routine maintenance and maintenance engineering interventions, the distribution of historical data is non-natural decay, and the frequency of once-a-year detection results in very few samples. The numerous factors have led to the lack of mature pavement performance prediction applications in the maintenance field.

In summary, the road performance prediction method is based on small sample data in the current situation. Except for the small sample, we also need to consider the situation of maintenance engineering intervention, thus, we prefer to construct the Long-short time memory (LSTM)\cite{LSTMppp} model in our proposed framework, because LSTM can remember information for the long term. This is done by remembering the previous output and combining it with the current one. This ability determines that LSTM can solve the nonlinear problem of data brought about by maintenance engineering intervention better than other models.
The proposed pavement performance prediction model is composed of two parts:

\begin{enumerate}
\item[1)]Use feature engineering to capture the relation between the pavement performance index and each disease, and use selected diseases as features to construct the LSTM model to predict the future number of each pavement disease, similar with some scholars use LSTM for pavement temperature predict\cite{Miland2021PavementTem}. 

\item[2)]Use all of the historical data to construct and train the multiple linear regression model, and predict the performance of the pavement using each future number of disease results of step 1).
\end{enumerate}

Assume $C_1, ~C_2, ~C_3$ as the features that are through the feature engineering cluster filtered. 
$C_1=\{c_1^{(1)},c_2^{(1)},...,c_t^{(1)}\}$, $C_2=\{c_1^{(2)},c_2^{(2)},...,c_t^{(2)}\}$, $C_3=\{c_1^{(3)},c_2^{(3)},...,c_t^{(3)}\}$, where $t$ denote to the size of the sample also represent the time. Then we need to use the multivariate LSTM model to predict the next 5 years' value of each feature. 

Once we have completed predictions for several diseases selected by feature engineering and next is training the multiple linear regression model, due to the large scale of the road network, the influencing factors are also intricate, so, it is better to build a model, and the pavement performance of all roads in the road network can be accurately predicted.
Assume that the pavement performance corresponds to the sequence of the three features data of $C_1,~C_2,~C_3$ is $Y = \{y_1,~ y_2, ...,y_t\}$, then the multiple linear regression model is defined as:
\begin{equation}
	y_{p} = \beta_0+\Sigma_{t=1}^n W_t c_t^{(i)}
\end{equation}

where $c_t^{(i)}$ denotes the number of diseases of the $i^{th}$ disease at the time $t$, $\beta_0$ and $W_t$ denotes the bias and the weight vector of $c_t^{(i)}$. Using our input data to train the model, we define the Sum of Squared Residuals(SSR) loss function, it can be described as:

\begin{equation}
	Loss = \Sigma_{t=1}^n(y_t-y_{p})^2
\end{equation}
SSR is one of the most common loss functions, where we sum up the squared of all the errors.

\subsection{Priority rank model}\label{RM}
The priority ranking model consists of two components: the maintenance route prioritization model and the maintenance segment prioritization model. Both models are built based on historical maintenance data.
\subsubsection{Prioritization of maintenance routes}\label{Prioritization of maintenance routes}
In general, as pavement performance decreases, maintenance costs will significantly increase. However, the social and economic benefits derived from the maintenance efforts also become more pronounced. Therefore, in practical maintenance project decision-making, segments with lower pavement performance typically have higher maintenance priorities. However, this is not an absolute rule. In the prioritization of maintenance routes, the maintenance probabilities of segments within the route should also be considered. By performing the necessary computations, the obtained probability values serve as the basis for prioritizing the routes. Higher values indicate higher priority levels. 

The pseudocode \ref{alg: assign Pro to data point} shows how to assign a probability to the predicted value of the pavement performance of each route.

\begin{algorithm}[h]
  \SetAlgoLined
  \KwData{Predicted value of pavement performance: [$x_1,x_2,\dots x_n$] $x_n$ represents the predicted value of route $n$ }
  \KwResult{Probabilities for each data point: [$P(x_1),P(x_2),\dots P(x_n)$]}
  \label{alg: assign Pro to data point}
  Sort the data points in ascending order\;
  Define the cumulative distribution function (CDF) as: $CDF(x) = (x - \text{min\_val}) / (\text{max\_val} - \text{min\_val})$\;
  Calculate the probability for each data point\;
  Initialize P(X) = []\;
  \For{each data point $x$ in the sorted data points}{
    \uIf{$x$ is the first data point}{
      probability($x$) = 1- (~CDF($x$) - 0~)\;
      P(X) append(probability($x$))\;
    }
    \Else{
      probability($x$) = 1- (~CDF($x$) - CDF(previous~data~point)~)\;
      P(X) append(probability($x$))\;
    }
  }
  \caption{Assigning probabilities to data points}
\end{algorithm} 

Therefore, the calculation method for prioritizing maintenance routes should be as follows:
\begin{equation}
Prioity~routes = Rank[P(x_n)*(1-P(x_{n,m}))]
\end{equation}
where $P(x_{n,m})$ denotes under the given conditions, the probability of road segment $m$ in route $n$ being assigned a maintenance project. How to calculate $P(x_{n,m})$ will be introduced in section \ref{Prioritization of maintenance road segments}

\subsubsection{Prioritization of maintenance road segments}\label{Prioritization of maintenance road segments}
\begin{figure}[h]
    \centering
    \includegraphics[width=2.5in]{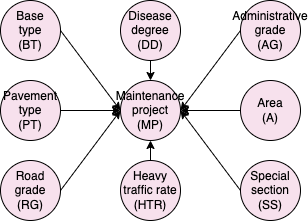}
    \caption{Bayes network to construct the rank model including both technical and non-technical considerations.}
    \label{fig: rankmodel}
\end{figure}

The road segments rank model is employed to determine the maintenance priorities for individual road segments. This model assigns a ranking or priority value to each road segment based on certain criteria or factors. In addition to the initial technical indicators. To enhance the applicability to regional maintenance management units and account for the impact of non-technical factors, additional considerations are incorporated beyond the initial technical indicators. Specifically, historical maintenance data is utilized to capture both management expertise and technical indicators. So, the Bayes network sees Fig \ref{fig: rankmodel} is employed as the primary model for prioritization, its inherent ability to capture and update probabilistic dependencies among variables aids in incorporating diverse factors, including both technical and non-technical considerations, into the decision-making process.

As Fig \ref{fig: rankmodel} shows, the maintenance project will be assigned or not depending on eight factors, including the technical and non-technical indicators, given those factors, $P(x_{n,m})$ can be calculated from the following formula:
\begin{equation}\label{rankbayes}
P(MP|BT, PT, RG, DD, HTR, AG, A, SS)
\end{equation}
through equation \ref{rankbayes} can get a list of the probability, the higher the probability under a given condition means the higher the priority of road segment maintenance under that condition. However, it is still necessary to consider major factors such as maintenance budget and maintenance effectiveness. Therefore, the algorithm for prioritizing maintenance sections can be described as pseudocode \ref{alg: generate priority seg list}:

\begin{algorithm}[!h]
  \SetAlgoLined
  \KwData{Data with factors: [$\text{Factor1}_1, \text{Factor2}_1, \dots, \text{FactorN}_1$], [$\text{Factor1}_2, \text{Factor2}_2, \dots, \text{FactorN}_2$], $\dots$, [$\text{Factor1}_M, \text{Factor2}_M, \dots, \text{FactorN}_M$]}
  \KwResult{Priority list of data points}
  \label{alg: generate priority seg list}
  \textbf{Step 1: Logistic Regression to Determine Weights}\;
  Combine the factors into a feature matrix $X$ and the decisions into a target vector $y$\;
  Create and fit a logistic regression model using $X$ and $y$\;
  Retrieve the estimated coefficients (weights) from the model\;
  
  \textbf{Step 2: Bayesian Optimization for Priority List}\;
  Define the scoring function using the weights obtained from Step 1\;
  Define the objective function for Bayesian optimization\;
  Define the search space for Bayesian optimization\;
  Run Bayesian optimization using the objective function and search space\;
  Retrieve the optimal data point based on the maximized objective\;
  
  \textbf{Step 3: Generate Priority List}\;
  Calculate scores for all data points using the scoring function\;
  Sort the data points in descending order based on their scores\;
  Create an empty priority list\;
  \For{each data point in the sorted data points}{
    Add the data point to the priority list
  }
  
  \textbf{Step 4: Output Priority List}\;
  Output the generated priority list
  
  \caption{Generating Priority List using Logistic Regression and Bayesian Optimization}
\end{algorithm}

\subsection{Road segment treatment recommendation model}\label{maintenance road segment recommendation model} 

\begin{figure*}[!t]
    \centering
    \includegraphics[width=4.5in]{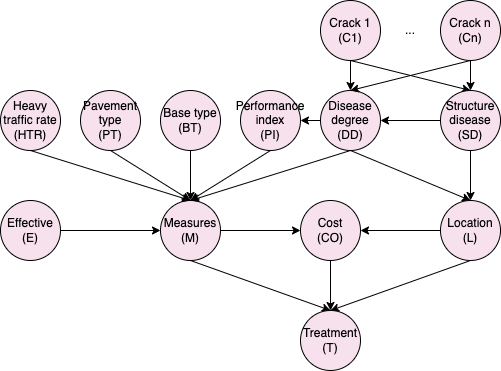}
    \caption{Bayes network to construct reward function in reinforcement learning, the given network shows the relation of each factor and can calculate the probability of each factor}
    \label{fig: bayesnet}
\end{figure*}
In general, as the table \ref{rpdddata} shows, pavement performance data include the index value, several kinds of cracks, and basic information about inspected road sections like pavement type et. Maintenance history records also include basic information about road sections and other detailed data on maintenance engineering design and implementation.

The most important part of reinforcement learning is to define the reward function $R(S, a, S')$, according to the "prediction policy problems" proposed by Kleinberg in 2015 \cite{KleinbergJon2015PPP}, the core solution can be described as follows:

\begin{equation}
\frac{d\pi (X_0,Y)}{dX_0} = \frac{\partial \pi}{\partial X_0}\ \underbrace{(Y)}_{\rm predict}\ + \frac{\partial \pi}{\partial Y} \underbrace{\frac{\partial Y}{\partial X_0}}_{\rm causation}
\end{equation}
where $Y$ be an outcome variable which depends in an unknown way on a set of variables $X_0$ and $X$.

Consider "predict policy" and combined with the real environment situation, the average cost of each maintenance policy per kilometer will be the key factor in the reward function, denote to $C(\pi(a))$ which can be searched from maintenance history records. To define the reward function we can use follows equation:

\begin{equation}
R(S, a, S') =  \frac{C(\pi(a))}{Y-\underbrace{y_{p}}_{\rm predict}} + \underbrace{\lambda_{a} C(\pi(a))}_{\rm Long-term \ evaluation}
\end{equation}

where $Y$ denote the pavement performance value after action $a$ is performed, $y_p$ denotes the prediction of the pavement performance in the next year, $d$ denotes the mileage of the segment to be maintained, and $D$ denotes the mileage of the road network. $\lambda_a$ represents the attenuation rate of the pavement performance index under the policy $a$ can calculate by:
\begin{equation}
\lambda _a =  \frac{1}{n}\sum_{i=1}^n Y_i^{(t)}-y_i^{(t+1)}
\end{equation}
where $n$ denotes the sample size for acting $a$, and $Y_i^{(t)}$ is an index means the value of $i^{th}$ sample pavement performance after acting $a$, $y_i^{(t+1)}$ means the pavement performance value in the next year after action $a$ is performed.

Based on our introduction to reinforcement learning in \ref{LR}, we also need the probability of each state, to calculate the probability of each state on the input data, we should rely on the given Figure \ref{fig: bayesnet}, the graph shows the Bayes network construct of each factor we need, the terminal node named effective that is our state will use in bellman equation to evaluate our maintenance policy.

As the given Bayes network and the given data, we can calculate the probability of the next state easily. Assume for a specific section of road, the expected road performance is increased $p$ points higher than the current one, we can calculate the probability of each possible measure based on given base data of road:
\begin{equation}\label{propability of measures}
P(M|HTR, PT, BT, PI, DD)
\end{equation}
Second, we need to calculate the probability of each possible location based on the disease degree.
\begin{equation}\label{Probability of location}
P(L|DD, SD)
\end{equation}
Over here, combining the equation \ref{propability of measures} and \ref{Probability of location}, we already know the probability of each treatment method:
\begin{equation}\label{PL} 
P(T|L, M, CO)
\end{equation}
The equation \ref{PL} is the probability of each possible action to improve road performance.

\section{Case Study}
In this experiment, the data used is actual monitoring data obtained from an automated road condition detection system in Xinjiang province of China, the detection data and evaluation data both satisfy the criteria specified in Standard "A". Due to the large volume of data, the experiment code and data can be accessed at \url{http://example.com}.

We validated the technical feasibility of the proposed intelligent decision-making framework using a small road network composed of three routes. The maintenance project implementation plan based on the 2021 data has already been released. According to the framework process proposed in this paper, we will formulate the 2021 highway maintenance plan based on the actual detection data from 2020. Subsequently, we will evaluate our framework by comparing it with the 2021 actual maintenance project implementation plan.
As table \ref{rpdddata} shows 2013-2021 automatic detection of pavement data on a small road network consisting of three routes, the PCI in the table is one of the indicators reflecting pavement performance in China.

\begin{table*}[!t] 
	\centering
	\caption{Pavement performance detect data}  
	\begin{tabular}{*{10}{c}}
		\hline\hline\noalign{\smallskip}	
		Road ID & Road grade & Pavement type & Base type & Traffic volume & Department & Unit & Area & Special section & Administrative grade \\
		\noalign{\smallskip}\hline\noalign{\smallskip}
		A000 & A & A & A & M & A00 & H000 & A & 1 & A \\
		C000 & A & A & A & L & C00 & P003 & A & 1 & B \\
		J000 & A & A & A & H & O00 & L003 & A & 0 & A \\
		\noalign{\smallskip}\hline
	\end{tabular}
	\label{rpdddata}
\end{table*}

\subsection{Pavement Performance Predict}

\begin{figure*}[!t]
    \centering
    \includegraphics[width=6.5 in]{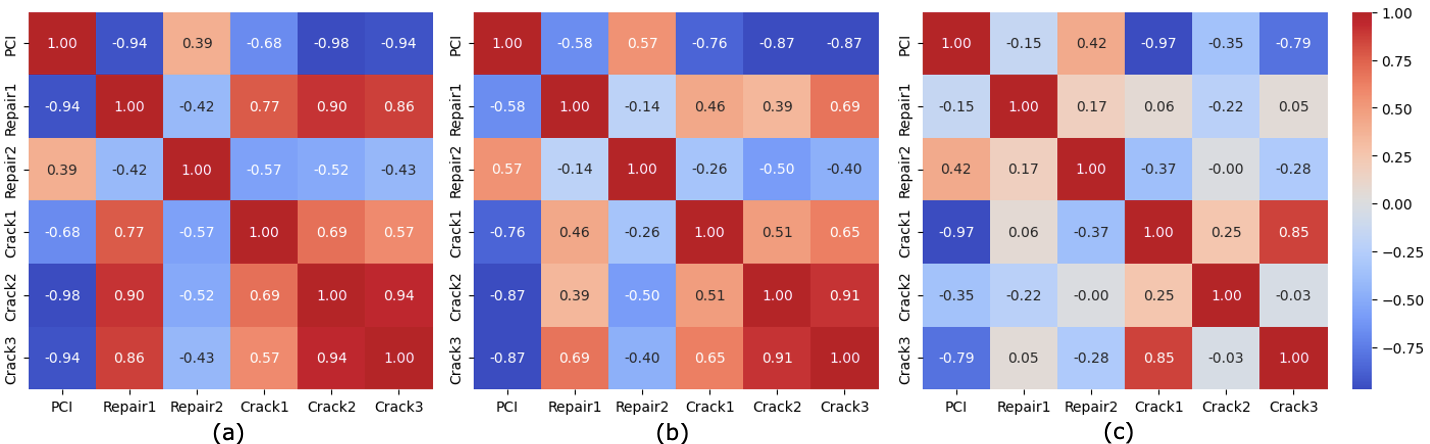}
    \caption{Figure (a) shows the correlation between the PCI value of road route 1 and the five types of cracks; Figure (b) shows the correlation between the PCI value of road route 2 and the five types of cracks; Figure (c) shows the correlation between the PCI value of road route 3 and the five types of cracks;}
    \label{fig:roadcorr}
\end{figure*}

In Section \ref{pavement performance prediction model}, we have provided a comprehensive overview of the construction of pavement performance prediction models. By analyzing the features of the three road segments presented in Table \ref{rpdddata}, we have identified the highly correlated crack features within each route, as illustrated in figure \ref{fig:roadcorr}, in our experiments, we selected pavement cracks with an absolute correlation coefficient greater than or equal to 0.7 as features.

Our experiment on this part used a learning rate of 0.01. Due to the small size of our data samples, the network structure was kept as simple as possible, resulting in a single hidden layer. The number of neurons in the hidden layer, which was set to 32, 64, or 128, was determined based on the distribution of different features in the dataset. For the size of the window, we suggest using 3, in certain exceptional cases, we still encourage adjusting this hyperparameter to other values. Additionally, the input data was divided considering the distribution of feature data.

From Table \ref{tab:corr}, we can observe significant discrepancies between the predicted results for 2021 and the actual detected road conditions in terms of cracks or PCI. However, by examining the data distribution in the table, we can notice that there is intervention from maintenance projects nearly every year. Therefore, the actual 2021 detection results do not reflect the natural deterioration of road conditions. In contrast, the predicted values are very similar to the degradation of pavement performance and disease development under natural conditions. Consequently, we can conclude that the LSTM + multiple linear regression model demonstrates strong predictive capabilities even without excluding the effects of maintenance interventions.

\begin{table*}[!t] 
	\centering
	\caption{Road Pavement Disease Predict result}  
	\begin{tabular}{*{10}{c}}
		\hline\hline\noalign{\smallskip}	
		Route ID & Year & PCI & Repair$_1$ & Repair$_2$ & Crack$_1$ & Crack$_2$ & Crack$_3$ \\
		\noalign{\smallskip}\hline\noalign{\smallskip}
		A000 & 2013 & 93.98 &0.81 &168.40 & 0 & 2.06 & 2.86 \\
		A000 & 2014 & 89.24 &2.78 &210.28 & 0.54 & 41.36 & 38.69  \\
		A000 & 2015 & 89.96 &7.52 &350.44 & 0 & 15.89 & 15.62 \\
		A000 & 2016 & 87.04 &2.55 &219.53 & 0.02 & 67.04 & 39.66 \\
		A000 & 2017 & 87.03 &2.93 &229.42 & 0.08 & 54.68 & 133.32\\
		A000 & 2018 & 78.55 &3.56 &80.60 & 1.40 & 269.43 & 240.38 \\
		A000 & 2019 & 78.18 &1.12 &148.17 & 0 & 239.67 & 183.85 \\
		A000 & 2020 & 78.86 &12.44 &192.26 & 2.36 & 206.82 & 184.51 \\
		A000 & 2021 & 73.83 &13.60 &57.18 & 8.94 & 331.76 & 244.14 \\
		\noalign{\smallskip}\hline\noalign{\smallskip}
		Predict Route A000 & 2021& 73.67& - & - & 4.552 & *295.67 & *355.49\\
		\noalign{\smallskip}\hline\noalign{\smallskip}
		C000 & 2013 & 72.75 &78.42	&566.26	& 84.35	& 191.38 & 123.25  \\
		C000 & 2014 & 76.63 & 7.62 & 228.53 & 11.85  & 333.20 & 99.94 \\
        C000 & 2015 & 94.52 & 4.87 & 1946.71 & 0 & 10.34 & 1.91 \\
        C000 & 2016 & 87.94 & 1.93 & 160.49 & 0.58 &  100.29 & 35.35 \\
        C000 & 2017 & 90.57 & 15.52 & 313.00 & 0.61 &  20.90 & 9.56 \\
        C000 & 2018 & 85.51 & 34.79 & 431.11 & 2.48 &  78.19 & 42.64 \\
        C000 & 2019 & 72.96 & 10.10 & 6.61 & 0.94  & 227.66 & 44.49 \\
        C000 & 2020 & 69.22 & 49.88 & 276.09 & 94.57  & 449.35 & 194.76 \\
        C000 & 2021 & 59.46 & 44.52 & 57.14 & 418.75  & 344.73 & 160.24 \\
        \noalign{\smallskip}\hline\noalign{\smallskip}
        Predict Route C000 & 2021& 52.87 & - & - & *94.57 & *409.84 & *194.57\\
        \noalign{\smallskip}\hline\noalign{\smallskip}
        J000 & 2013 & 89.47 & 0.94 & 8.28 & 7.19 & 17.75 & 24.85\\
        J000 & 2014 & 70.80 & 0.54 & 5.34 & 93.11 & 101.95 & 75.63\\
        J000 & 2015 & 78.24 & 2.55 & 482.63 & 52.26 & 127.93 & 98.72\\
        J000 & 2016 & 59.37 & 2.37 & 38.89 & 325.29 & 178.85 & 122.28\\
        J000 & 2017 & 44.96 & 4.37 & 37.40 & 671.31 & 198.55 & 123.87\\
        J000 & 2018 & 28.81 & 1.41 & 16.43 & 1412.00 & 190.82 & 125.32\\
        J000 & 2019 & 29.95 & 0.00 & 23.05 & 1301.29 & 245.59 & 1306.75\\
        J000 & 2020 & 14.87 & 0.70 & 4.36 & 1818.37 & 176.10 & 729.65\\
        J000 & 2021 & 13.14 & 4.38 & 0.33 & 1902.71 & 27.69 & 8.80\\
        \noalign{\smallskip}\hline\noalign{\smallskip}
        Predict Route J000 & 2021& 0 & - & - & *2354.60 & 897.48 & *1937.21\\
		\noalign{\smallskip}\hline
	\end{tabular}
	\label{tab:corr}
	\begin{flushleft}
        \textit{*}: The predicted values preceded by the symbol "*" denote that this predicted value is one of the features in this route and they will be used in the multiple linear regression modeling and predictions.
    \end{flushleft}
\end{table*}

\subsection{Prioritization of maintenance routes.}

\begin{table}[h]
	\centering
	\caption{Prioritization of maintenance routes}  
	\begin{tabular}{*{5}{c}}
		\hline\hline\noalign{\smallskip}	
		Route ID & Pre-PCI & $P(x_n)$ & $P(x_{n,m})$ & Prioritization of routes \\
		\noalign{\smallskip}\hline\noalign{\smallskip}
		A000 & 73.67 & 0 & 0.36 & 0 \\
		C000 & 52.87 & 0.72 & 0.13 & 0.63 \\
		J000 & 0.00 & 1 & 0.21 & 0.79 \\
		\noalign{\smallskip}\hline
	\end{tabular}
	\label{tab: Prioritization of maintenance routes}	
\end{table}

In section \ref{RM}, we have provided a comprehensive overview of the rank model, as the table \ref{tab: Prioritization of maintenance routes} shows, based on the provided one-year pavement performance prediction values and maintenance history data, and under a certain maintenance budget constraint, it is determined that the priority order of the three routes in the experiment should be Route J000 followed by Route COOO and then Route A000.

\subsection{Maintenance Recommendation Plan and Road Section Prioritization}

\begin{figure}[thbp]
    \centering
    \includegraphics[width=3 in]{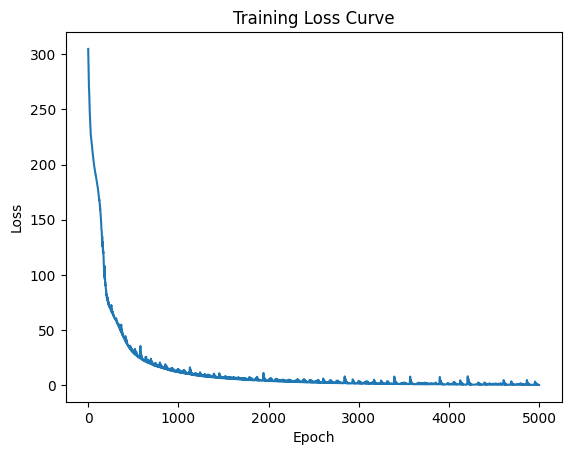}
    \caption{The training loss function drops rapidly within 1000 epochs, and the subsequent decline is slower, and tends to be stable around the 5000th epoch;}
    \label{fig: Deeplrloss}
\end{figure}

In this part, the parameter settings in deep Q-learning, such as the learning rate and decay rate, were referenced from \cite{RLinHighway}. Experimental results have also demonstrated that these parameter settings are optimal. Additionally, other parameter settings are presented in the table \ref{tab: deep_q_para}:

Figure \ref{fig: Deeplrloss} depicts the learning curve of the neural network, which clearly indicates that the model reaches a stable state around the 3000th epoch. This observation suggests that, under the influence of multiple factors, the model has achieved an optimal state for the recommendation strategy using the reward function as the evaluation metric. Further optimization beyond this point appears unattainable.

\begin{figure}[thbp]
    \centering
    \includegraphics[width=3 in]{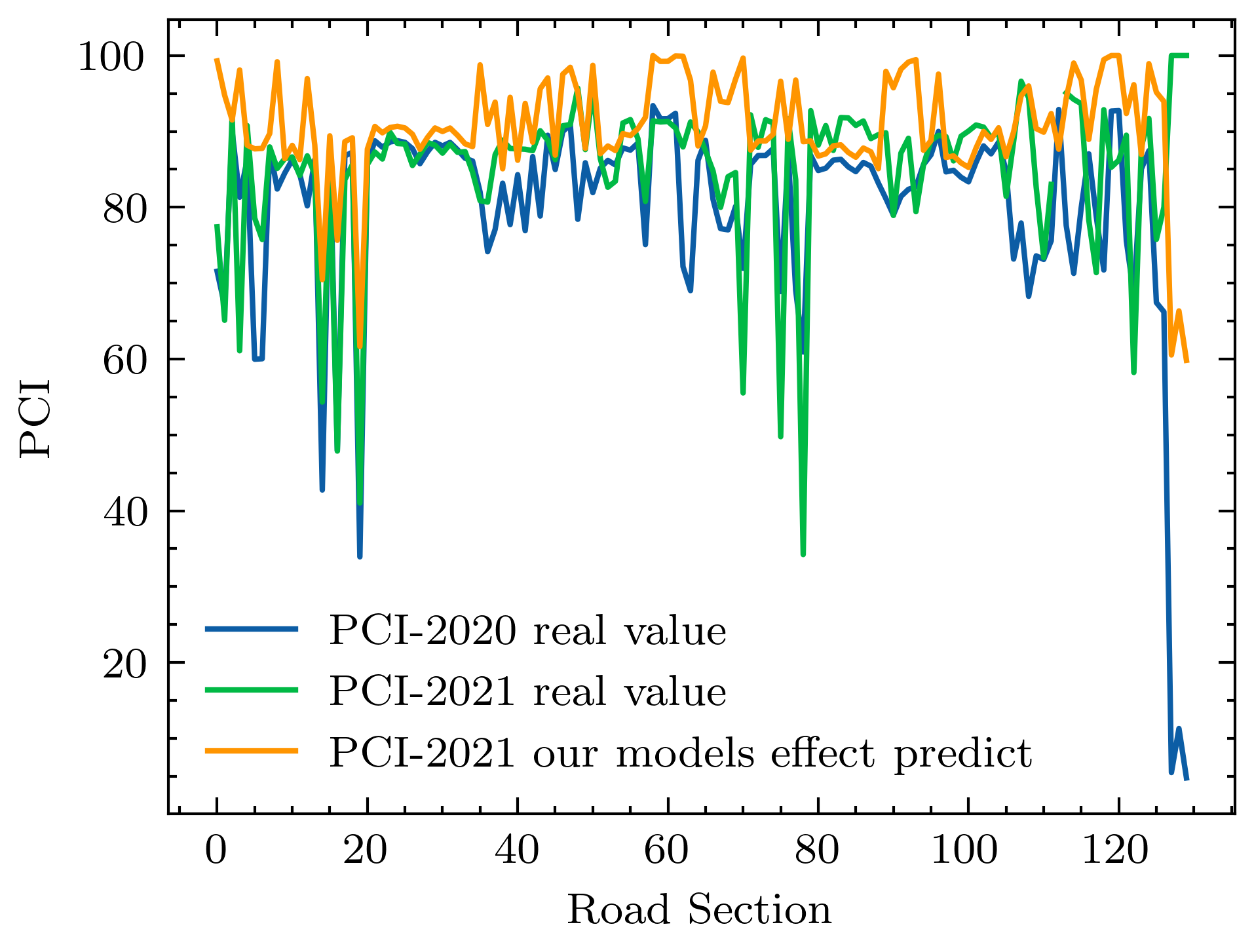}
    \caption{A000 Prioritization of Road Sections in Maintenance Plan and Maintenance Effectiveness Line Graph;}
    \label{fig: G217_treatment}
\end{figure}

\begin{figure}[thbp]
    \centering
    \includegraphics[width=3 in]{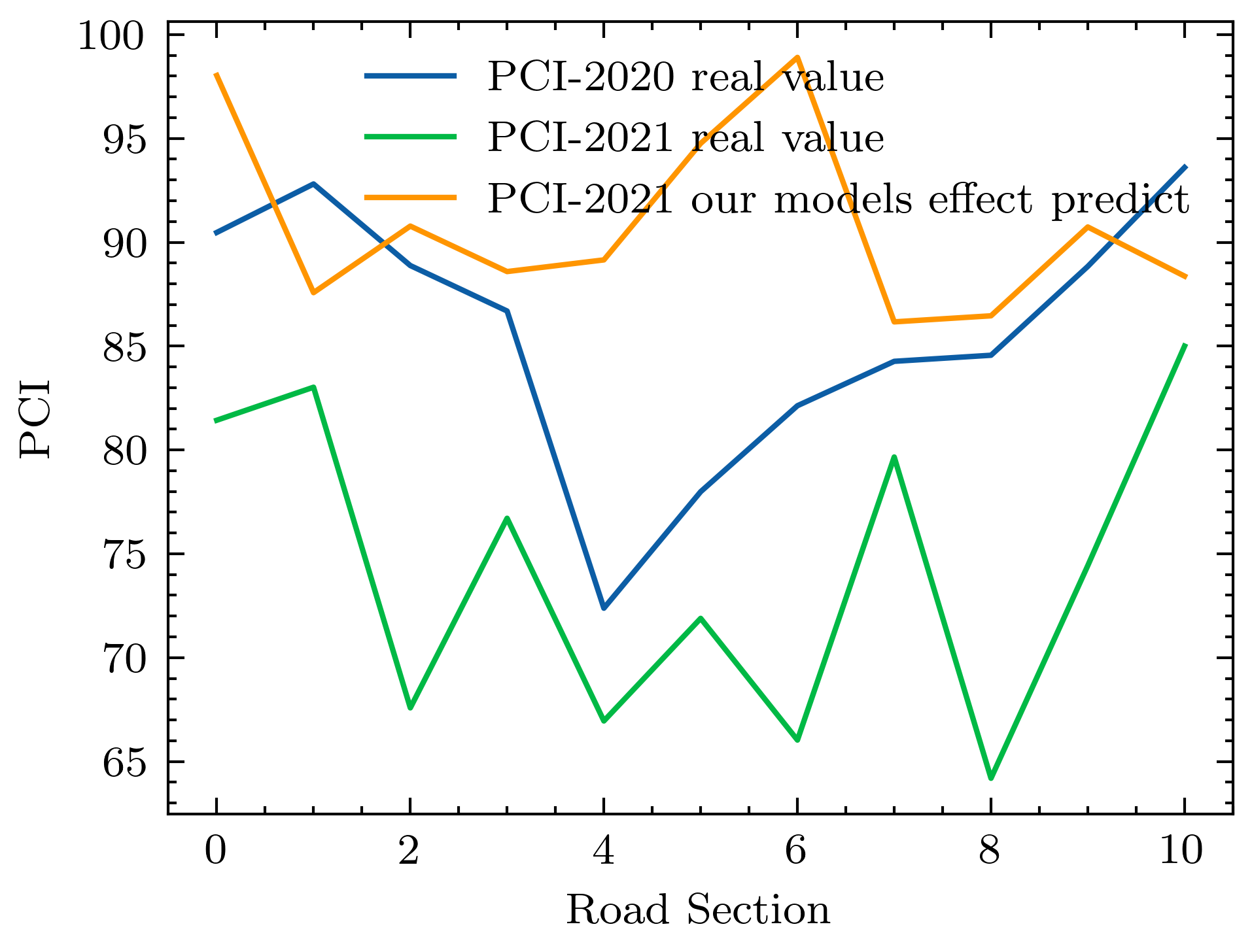}
    \caption{C000 Prioritization of Road Sections in Maintenance Plan and Maintenance Effectiveness Line Graph;}
    \label{fig: G218_treatment}
\end{figure}

\begin{figure}[thbp]
    \centering
    \includegraphics[width=3 in]{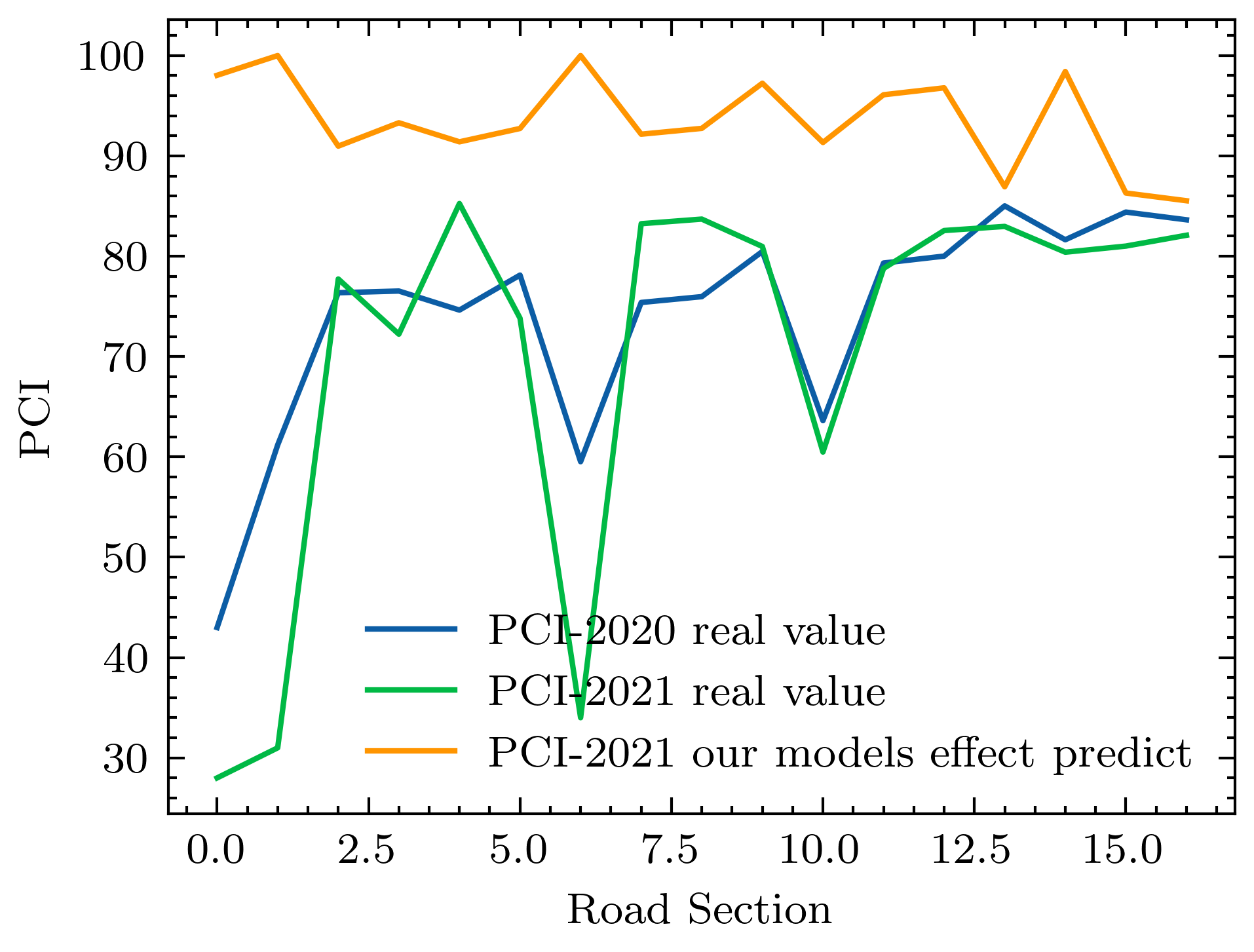}
    \caption{J000 Prioritization of Road Sections in Maintenance Plan and Maintenance Effectiveness Line Graph;}
    \label{fig: G312_treatment}
\end{figure}

Figure \ref{fig: G217_treatment}, \ref{fig: G218_treatment}, and \ref{fig: G312_treatment} illustrates the predicted effectiveness curves of maintenance recommendations for three routes, namely A000, C000, J000, generated by our proposed intelligent decision-making framework, in the graph, the horizontal axis represents the priority of road sections in ascending order. The smaller the index of a road section, the higher its priority. It is important to note that this index does not indicate the starting coordinates of the road sections. To determine the specific road sections, one should refer to the original data of each route and retrieve the road sections based on their corresponding indices, thereby identifying each section's start and endpoints. 

The curves also include the actual results of the maintenance plans based on automated condition monitoring in 2020 and the maintenance plans formulated by the management unit with their expertise in 2021. From the graph, it is evident that the maintenance recommendations provided by the model consistently outperform the empirically developed maintenance plans. While further validation through the real-world application is necessary to confirm the actual effectiveness, it is worth noting that the effectiveness of the recommended maintenance plans is based on historical evaluations of maintenance plan outcomes. Therefore, representing the evaluation data, the red curve closely approximates the actual maintenance effectiveness. In addition, by examining these three graphs, it can be observed that the fluctuations in the red curve, resulting from the implementation of functional repair-based maintenance plans, differ from the patterns exhibited by the other two curves. However, in the case of preventive maintenance plans, the majority of road condition fluctuations align with those of the other two actual condition curves. This observation provides further evidence that the recommended maintenance plans generated by the model closely approximate the actual maintenance effectiveness, thus validating the feasibility of the maintenance plan recommendation model. Additionally, since the red curve mostly lies above the other two curves, it further confirms the optimality of the recommended maintenance plans suggested by the model.

\begin{table}[h] 
	\centering
	\caption{Parameter List for Reinforcement Learning}  
	\begin{tabular}{*{2}{c}}
		\hline\hline\noalign{\smallskip}	
		Parameters & Values of parameters\\
		\noalign{\smallskip}\hline\noalign{\smallskip}
		Number of Layers & 5  \\
		Number of hidden layers & 3 \\
		Total parameters & 75589 \\
		Discount factors & 0.9\\	
		Learning rate & 0.001\\
		Number of epochs & 5000\\
		Start year & 2019\\
		End Year & 2020\\
		\noalign{\smallskip}\hline
	\end{tabular}
	\label{tab: deep_q_para}
\end{table}

\section{Conclusion}
In order to formulate an optimal decision for highway pavement maintenance planning, various factors need to be considered. From a management perspective, these factors include administrative hierarchy, highway technical classification, traffic volume, maintenance budget, and even natural environmental factors. Furthermore, from a technical standpoint, the analysis should encompass not only the current pavement condition but also factors such as the structural integrity of the highways.

Therefore, this paper constructs a decision-making environment that incorporates multiple factors. It combines the current and predicted future road conditions with historical evaluations of maintenance plan effectiveness to establish a reward function. By employing reinforcement learning, the study achieves optimization decisions that closely approximate real-world engineering applications.

The modeling process and analysis of decision outcomes are illustrated through a case study. Several conclusions can be drawn regarding the developed Deep Reinforcement Learning (DRL) method:

\begin{enumerate}
\item[1)]The deep reinforcement learning model proposed in this paper greatly assists decision-makers in determining the optimal maintenance plans. These maintenance plans are derived through a comprehensive evaluation of the combined effects of 60 different historical maintenance plans, 55 base thicknesses, 41 pavement widths, and 17 pavement thicknesses. This approach enables the identification of the optimal maintenance plan decisions for varying road conditions, thereby achieving effective decision-making for different scenarios.

\item[2)]By incorporating future predictions into deep reinforcement learning, the model can accurately simulate the deterioration of road conditions under different scenarios. This enhanced capability allows for a more precise determination of effective maintenance plans. By considering the anticipated changes in road conditions, decision-makers can proactively identify and implement appropriate maintenance strategies to mitigate potential issues and ensure optimal road maintenance outcomes.
\item[3)]Based on the evaluation of the expected effectiveness of maintenance plans recommended by the model, it is evident that the recommended plans are indeed more effective than the decisions made by maintenance managers relying solely on empirical models. This finding highlights the superiority of the model-driven approach in optimizing maintenance decision-making processes. 
\end{enumerate}

The model proposed in this paper focuses on achieving the optimal effectiveness for each maintenance plan decision, without explicitly considering the maximum budget for each road route. However, after determining the maintenance plans, we employ the Bayesian optimization algorithm to establish the prioritization of road section maintenance. Managers can then allocate budgets for each road route based on the prioritization of maintenance for each section. By considering the maintenance budget for each route and the prioritization of road section maintenance, the managers can prioritize the maintenance of high-priority road sections within the constraints of the budget allocation. This approach ensures that road sections with higher priority receive preferential maintenance treatment while adhering to the limitations imposed by the budget constraints.

\section*{Thanks} 

Here, we would like to thank Dr. Guo Yuanhao, Dr. Pan Zongjun, and Technical Director Wang Yuqiang of RoadMainT Co., Ltd. for their suggestions and feedback on this study. At the same time, we would also like to thank Engineer Liu Yingzhou of the Xinjiang Autonomous Region Highway Administration Bureau for his contribution to the engineering verification of the framework of this article.

\printbibliography

@Article{RLintran,
AUTHOR = {Xiang, Xuanchen and Foo, Simon and Zang, Huanyu},
TITLE = {Recent Advances in Deep Reinforcement Learning Applications for Solving Partially Observable Markov Decision Processes (POMDP) Problems Part 2—Applications in Transportation, Industries, Communications and Networking and More Topics},
JOURNAL = {Machine Learning and Knowledge Extraction},
VOLUME = {3},
YEAR = {2021},
NUMBER = {4},
PAGES = {863--878},
URL = {https://www.mdpi.com/2504-4990/3/4/43},
ISSN = {2504-4990},
DOI = {10.3390/make3040043}
}

@article{LTPP,
author = {Xueqin Chen  and Hehua Zhu  and Qiao Dong  and Baoshan Huang },
title = {Optimal Thresholds for Pavement Preventive Maintenance Treatments Using LTPP Data},
journal = {Journal of Transportation Engineering, Part A: Systems},
volume = {143},
number = {6},
pages = {04017018},
year = {2017},
doi = {10.1061/JTEPBS.0000044},

URL = {https://ascelibrary.org/doi/abs/10.1061/JTEPBS.0000044},
eprint = {https://ascelibrary.org/doi/pdf/10.1061/JTEPBS.0000044}
}

@article{priority,
    author = {Siswanto, Henri and Supriyanto, Bambang and Pranoto and Putra, Yusuf Akbar Megy and Huda, Alfian Syahrul},
    title = {Evaluation of road maintenance priority using PCI and road note 1 for Indonesian district roads},
    journal = {AIP Conference Proceedings},
    volume = {1977},
    number = {1},
    year = {2018},
    month = {06},
    issn = {0094-243X},
    doi = {10.1063/1.5042990},
    url = {https://doi.org/10.1063/1.5042990},
    note = {040020},
    eprint = {https://pubs.aip.org/aip/acp/article-pdf/doi/10.1063/1.5042990/14007822/040020\_1\_online.pdf},
}

@article{pfmnspvmntpre,
Author = {Bianchini, Alessandra and Bandini, Paola},
ISSN = {10939687},
Journal = {Computer-Aided Civil \& Infrastructure Engineering},
Number = {1},
Pages = {39 - 54},
Title = {Prediction of Pavement Performance through Neuro-Fuzzy Reasoning.},
Volume = {25},
URL = {https://ezproxy.gl.iit.edu/login?url=https://search.ebscohost.com/login.aspx?direct=true&db=a9h&AN=46766725&site=ehost-live},
Year = {2010},
}

@article{MLprepfmspvmt,
Author = {Marcelino, Pedro and de Lurdes Antunes, Maria and Fortunato, Eduardo and Gomes, Marta Castilho},
ISSN = {10298436},
Journal = {International Journal of Pavement Engineering},
Number = {3},
Pages = {341 - 354},
Title = {Machine learning approach for pavement performance prediction.},
Volume = {22},
URL = {https://ezproxy.gl.iit.edu/login?url=https://search.ebscohost.com/login.aspx?direct=true&db=a9h&AN=148772697&site=ehost-live},
Year = {2021},
}

@article{Fuzzyreg4pvmt,
title = {Pavement performance prediction through fuzzy regression},
journal = {Expert Systems with Applications},
volume = {38},
number = {8},
pages = {10010-10017},
year = {2011},
issn = {0957-4174},
doi = {https://doi.org/10.1016/j.eswa.2011.02.007},
url = {https://www.sciencedirect.com/science/article/pii/S0957417411002132},
author = {Nang-Fei Pan and Chien-Ho Ko and Ming-Der Yang and Kai-Chun Hsu}
}

@Article{MLPPPs,
AUTHOR = {Justo-Silva, Rita and Ferreira, Adelino and Flintsch, Gerardo},
TITLE = {Review on Machine Learning Techniques for Developing Pavement Performance Prediction Models},
JOURNAL = {Sustainability},
VOLUME = {13},
YEAR = {2021},
NUMBER = {9},
ARTICLE-NUMBER = {5248},
URL = {https://www.mdpi.com/2071-1050/13/9/5248},
ISSN = {2071-1050},
DOI = {10.3390/su13095248}
}

@ARTICLE{fuzzyLRanal,
  author={},
  journal={IEEE Transactions on Systems, Man, and Cybernetics}, 
  title={Linear Regression Analysis with Fuzzy Model}, 
  year={1982},
  volume={12},
  number={6},
  pages={903-907},
  doi={10.1109/TSMC.1982.4308925}}

@article{RLinCV,
author = {Le, Ngan and Rathour, Vidhiwar Singh and Yamazaki, Kashu and Luu, Khoa and Savvides, Marios},
address = {Dordrecht},
copyright = {The Author(s), under exclusive licence to Springer Nature B.V. 2021},
issn = {0269-2821},
journal = {The Artificial intelligence review},
language = {eng},
number = {4},
pages = {2733-2819},
publisher = {Springer Netherlands},
title = {Deep reinforcement learning in computer vision: a comprehensive survey},
volume = {55},
year = {2022},
}

@article{RLinRobotic,
author = {Singh, Bharat and Kumar, Rajesh and Singh, Vinay Pratap},
address = {Dordrecht},
copyright = {The Author(s), under exclusive licence to Springer Nature B.V. 2021},
issn = {0269-2821},
journal = {The Artificial intelligence review},
language = {eng},
number = {2},
pages = {945-990},
publisher = {Springer Netherlands},
title = {Reinforcement learning in robotic applications: a comprehensive survey},
volume = {55},
year = {2022},
}

@article{PPP,
author = {Kleinberg, Jon and Ludwig, Jens and Mullainathan, Sendhil and Obermeyer, Ziad},
address = {United States},
copyright = {Copyright 2015 American Economic Association},
issn = {0002-8282},
journal = {The American economic review},
language = {eng},
number = {5},
pages = {491-495},
publisher = {American Economic Association},
title = {Prediction Policy Problems},
volume = {105},
year = {2015},
}

@article{Markovppp,
Author = {Moreira, Andre V. and Tinoco, Joaquim and Oliveira, Joel R. M. and Santos, Adriana},
ISSN = {10298436},
Journal = {International Journal of Pavement Engineering},
Number = {10},
Pages = {937 - 948},
Title = {An application of Markov chains to predict the evolution of performance indicators based on pavement historical data.},
Volume = {19},
Year = {2018}
}

@article{ATTOHOKINE1999291,
title = {Analysis of learning rate and momentum term in backpropagation neural network algorithm trained to predict pavement performance},
journal = {Advances in Engineering Software},
volume = {30},
number = {4},
pages = {291-302},
year = {1999},
issn = {0965-9978},
doi = {https://doi.org/10.1016/S0965-9978(98)00071-4},
url = {https://www.sciencedirect.com/science/article/pii/S0965997898000714},
author = {Nii O. Attoh-Okine},
}

@misc{LSTMppp,
      title={Time Series Forecasting Using LSTM Networks: A Symbolic Approach}, 
      author={Steven Elsworth and Stefan Guttel},
      year={2020},
      eprint={2003.05672},
      archivePrefix={arXiv},
      primaryClass={cs.LG}
}

@article{RLinrdrc,
author = {Fan, Xudong and Zhang, Xijin and Wang, Xiaowei and Yu, Xiong},
address = {Cham},
copyright = {The Author(s) 2023},
issn = {2662-2521},
journal = {Journal of infrastructure preservation and resilience},
language = {eng},
number = {1},
pages = {8-19},
publisher = {Springer International Publishing},
title = {A deep reinforcement learning model for resilient road network recovery under earthquake or flooding hazards},
volume = {4},
year = {2023},
}

@article{RLinHighway,
author = {Yao, Linyi and Leng, Zhen and Jiang, Jiwang and Ni, Fujian},
issn = {1524-9050},
journal = {IEEE transactions on intelligent transportation systems},
language = {eng},
number = {11},
pages = {1-12},
publisher = {IEEE},
title = {Large-Scale Maintenance and Rehabilitation Optimization for Multi-Lane Highway Asphalt Pavement: A Reinforcement Learning Approach},
volume = {23},
year = {2022},
}

@article{KleinbergJon2015PPP,
author = {Kleinberg, Jon and Ludwig, Jens and Mullainathan, Sendhil and Obermeyer, Ziad},
address = {United States},
copyright = {Copyright© 2015 American Economic Association},
issn = {0002-8282},
journal = {The American economic review},
language = {eng},
number = {5},
pages = {491-495},
publisher = {American Economic Association},
title = {Prediction Policy Problems},
volume = {105},
year = {2015},
}

@article{RLinDataMing,
author = {Ahmed, Usman and Lin, Jerry Chun-Wei and Srivastava, Gautam and Chen, Hsing-Chung},
address = {Amsterdam},
copyright = {Copyright IOS Press BV 2022},
issn = {1064-1246},
journal = {Journal of intelligent \& fuzzy systems},
language = {eng},
number = {5},
pages = {4751-4758},
publisher = {IOS Press BV},
title = {Deep active reinforcement learning for privacy preserve data mining in 5G environments},
volume = {42},
year = {2022},
}

@ARTICLE{Miland2021PavementTem,
  author={Milad, Abdalrhman and Adwan, Ibrahim and Majeed, Sayf A. and Yusoff, Nur Izzi Md and Al-Ansari, Nadhir and Yaseen, Zaher Mundher},
  journal={IEEE Access}, 
  title={Emerging Technologies of Deep Learning Models Development for Pavement Temperature Prediction}, 
  year={2021},
  volume={9},
  number={},
  pages={23840-23849},
  doi={10.1109/ACCESS.2021.3056568}}
\end{document}